\documentclass[11pt,a4paper]{article}
\usepackage[hyperref]{conll-2019}
\usepackage{times}
\usepackage{latexsym}

\usepackage{url}

\usepackage{multirow}
\usepackage{graphicx}
\usepackage{enumitem}
\usepackage{amsmath}
\usepackage{amsfonts}
\usepackage{color}
\usepackage{bm}
\usepackage{verbatim}
\aclfinalcopy 

\title{Learning to Detect Opinion Snippet for Aspect-Based Sentiment Analysis}

\author{Mengting Hu\textsuperscript{1}\thanks{\; Work performed while interning at IBM Research - China.}  \quad\;\; Shiwan Zhao\textsuperscript{2}\thanks{\; Corresponding author.} \quad {\bf Honglei Guo\textsuperscript{2} \quad Renhong Cheng\textsuperscript{1} \quad Zhong Su\textsuperscript{2}} \\
\textsuperscript{1} Nankai University \quad \textsuperscript{2} IBM Research - China \\
mthu@mail.nankai.edu.cn, \{zhaosw, guohl\}@cn.ibm.com \\ chengrh@nankai.edu.cn, suzhong@cn.ibm.com
}

\date{}

\begin{document}
\maketitle
\begin{abstract}
Aspect-based sentiment analysis (ABSA) is to predict the sentiment polarity towards a particular aspect in a sentence. Recently, this task has been widely addressed by the neural attention mechanism, which computes attention weights to softly select words for generating aspect-specific sentence representations. The attention is expected to concentrate on opinion words for accurate sentiment prediction. However, attention is prone to be distracted by noisy or misleading words, or opinion words from other aspects. In this paper, we propose an alternative hard-selection approach, which determines the start and end positions of the opinion snippet, and selects the words between these two positions for sentiment prediction. Specifically, we learn deep associations between the sentence and aspect, and the long-term dependencies within the sentence by leveraging the pre-trained BERT model. We further detect the opinion snippet by self-critical reinforcement learning. Especially, experimental results demonstrate the effectiveness of our method and prove that our hard-selection approach outperforms soft-selection approaches when handling multi-aspect sentences. 
\end{abstract}

\section{Introduction}
Aspect-based sentiment analysis \cite{Pang:2008:OMS:1454711.1454712,liu2012sentiment} is a fine-grained sentiment analysis task which has gained much attention from research and industries. It aims at predicting the sentiment polarity of a particular aspect of the text. With the rapid development of deep learning, this task has been widely addressed by attention-based neural networks \cite{wang2016attention,Ma2017Interactive,cheng2017aspect,tay2018learning,wang2018clause}. To name a few, \newcite{wang2016attention} learn to attend on different parts of the sentence given different aspects, then generates aspect-specific sentence representations for sentiment prediction. \newcite{tay2018learning} learn to attend on correct words based on associative relationships between sentence words and a given aspect. These attention-based methods have brought the ABSA task remarkable performance improvement.

\begin{figure}
\centering
	\includegraphics[width=0.47\textwidth]{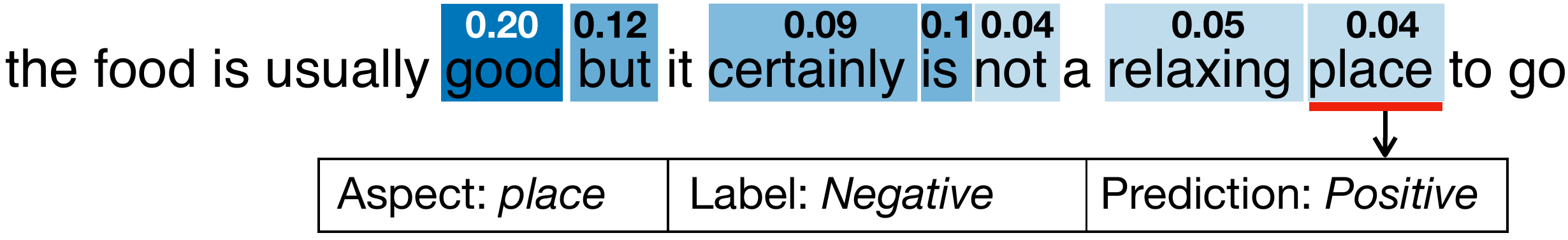}
    \caption{Example of attention visualization. The attention weights of the aspect \emph{place} are from the model ATAE-LSTM \protect\cite{wang2016attention}, a typical attention mechanism used for soft-selection.}
    \label{figure:example}
\end{figure}

Previous attention-based methods can be categorized as {\bf soft-selection} approaches since the attention weights scatter across the whole sentence and every word is taken into consideration with different weights. This usually results in attention distraction \cite{li2018acl}, i.e., attending on noisy or misleading words, or opinion words from other aspects. Take Figure \ref{figure:example} as an example, for the aspect \emph{place} in the sentence \emph{``the food is usually good but it certainly is not a relaxing place to go''}, we visualize the attention weights from the model ATAE-LSTM \cite{wang2016attention}. As we can see, the words \emph{``good''} and \emph{``but''} are dominant in attention weights. However, \emph{``good''} is used to describe the aspect \emph{food} rather than \emph{place}, \emph{``but''} is not so related to \emph{place} either. The true opinion snippet \emph{``certainly is not a relaxing place''} receives low attention weights, leading to the wrong prediction towards the aspect \emph{place}.

Therefore, we propose an alternative {\bf hard-selection} approach by determining two positions in the sentence and selecting words between these two positions as the opinion expression of a given aspect. This is also based on the observation that opinion words of a given aspect are usually distributed consecutively as a snippet \cite{wang2018learning}.
As a consecutive whole, the opinion snippet may gain enough attention weights, avoid being distracted by other noisy or misleading words, or distant opinion words from other aspects. We then predict the sentiment polarity of the given aspect based on the average of the extracted opinion snippet. The explicit selection of the opinion snippet also brings us another advantage that it can serve as justifications of our sentiment predictions, making our model more interpretable.

To accurately determine the two positions of the opinion snippet of a particular aspect, we first model the deep associations between the sentence and aspect, and the long-term dependencies within the sentence by BERT \cite{Devlin2018BERT}, which is a pre-trained language model and achieves exciting results in many natural language tasks. Second, with the contextual representations from BERT, the two positions are sequentially determined by self-critical reinforcement learning. The reason for using reinforcement learning is that we do not have the ground-truth positions of the opinion snippet, but only the polarity of the corresponding aspect. Then the extracted opinion snippet is used for sentiment classification. The details are described in the model section.

The main contributions of our paper are as follows: 
\begin{itemize}
\item We propose a hard-selection approach to address the ABSA task. Specifically, our method determines two positions in the sentence to detect the opinion snippet towards a particular aspect, and then uses the framed content for sentiment classification. Our approach can alleviate the attention distraction problem in previous soft-selection approaches.

\item We model deep associations between the sentence and aspect, and the long-term dependencies within the sentence by BERT. We then learn to detect the opinion snippet by self-critical reinforcement learning.

\item The experimental results demonstrate the effectiveness of our method and also our approach significantly outperforms soft-selection approaches on handling multi-aspect sentences.
\end{itemize}

\section{Related Work}
\label{sec:related}
Traditional machine learning methods for aspect-based sentiment analysis focus on extracting a set of features to train sentiment classifiers \cite{ding2009entity,boiy2009machine,jiang2011target}, which usually are labor intensive. With the development of deep learning technologies, neural attention mechanism \cite{Bahdanau2014Neural} has been widely adopted to address this task \cite{tang2015target,wang2016attention,tang2016aspect,Ma2017Interactive,chen2017recurrent,cheng2017aspect,li2018hierarchical,wang2018clause,tay2018learning,hazarika2018modeling,majumder2018iarm,fan2018multi,wang2018target}. \newcite{wang2016attention} propose attention-based LSTM networks which attend on different parts of the sentence for different aspects. \newcite{Ma2017Interactive} utilize the interactive attention to capture the deep associations between the sentence and the aspect. Hierarchical models \cite{cheng2017aspect,li2018hierarchical,wang2018clause} are also employed to capture multiple levels of emotional expression for more accurate prediction, as the complexity of sentence structure and semantic diversity. \newcite{tay2018learning} learn to attend based on associative
relationships between sentence words and aspect. 

All these methods use normalized attention weights to softly select words for generating aspect-specific sentence representations, while the attention weights scatter across the whole sentence and can easily result in attention distraction. \newcite{wang2018learning} propose a hard-selection method to learn segmentation attention which can effectively capture the structural dependencies between the target and the sentiment expressions with a linear-chain conditional random field (CRF) layer. However, it can only address aspect-term level sentiment prediction which requires annotations for aspect terms. Compared with it, our method can handle both aspect-term level and aspect-category level sentiment prediction by detecting the opinion snippet. 

\begin{figure}
\centering
	\includegraphics[width=0.47\textwidth]{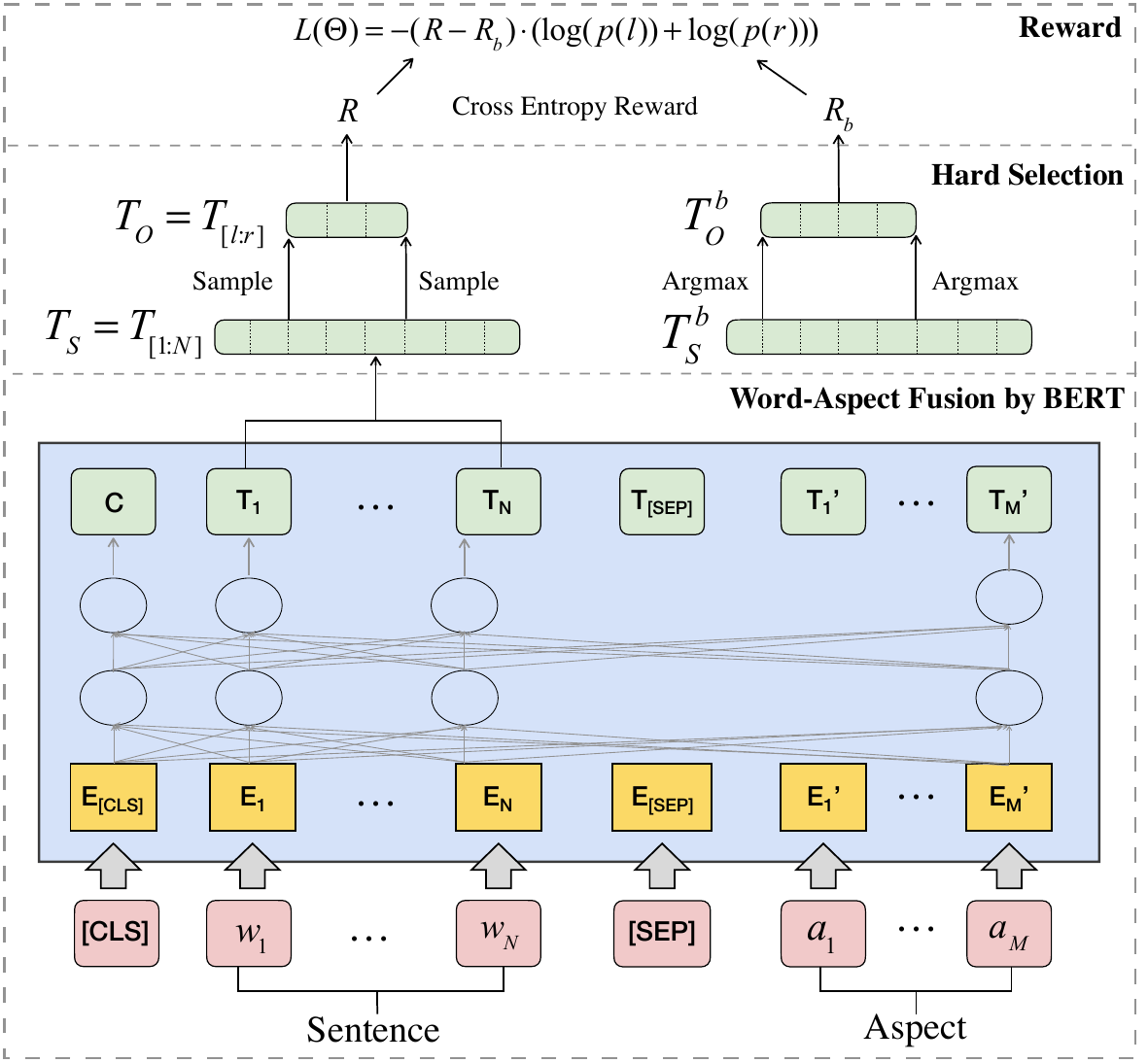}
    \caption{Network Architecture. We leverage BERT to model the relationships between sentence words and a particular aspect.  The sentence and aspect are packed together into a single sequence and fed into BERT, in which $E$ represents the input embedding, and $T_i$ represents the contextual representation of token $i$. With the contextual representations from BERT, the start and end positions are sequentially sampled and then the framed content is used for sentiment prediction. Reinforcement learning is adopted for solving the non-differentiable problem of sampling.}
    \label{network}
\end{figure}

\section{Model}
\label{sec:model}

We first formulate the problem. Given a sentence $S=\{w_1,w_2,...,w_N\}$ and an aspect $A=\{a_1,a_2,...,a_M\}$, the ABSA task is to predict the sentiment of $A$. In our setting, the aspect can be either aspect terms or an aspect category. As aspect terms, $A$ is a snippet of words in $S$, i.e., a sub-sequence of the sentence, while as an aspect category, $A$ represents a semantic category with $M=1$, containing just an abstract token.

In this paper, we propose a hard-selection approach to solve the ABSA task. Specifically, we first learn to detect the corresponding opinion snippet $O=\{w_{l},w_{l+1}...,w_{r}\}$, where $1\leq l\leq r\leq N$, and then use $O$ to predict the sentiment of the given aspect. The network architecture is shown in Figure \ref{network}.

\subsection{Word-Aspect Fusion}
Accurately modeling the relationships between sentence words and an aspect is the key to the success of the ABSA task. Many methods have been developed to model word-aspect relationships. \newcite{wang2016attention} simply concatenate the aspect embedding with the input word embeddings and sentence hidden representations for computing aspect-specific attention weights. \newcite{Ma2017Interactive}
learn the aspect and sentence interactively by using two attention networks. \newcite{tay2018learning}
adopt circular convolution of vectors for performing the word-aspect fusion.

In this paper, we employ BERT \cite{Devlin2018BERT} to model the deep associations between the sentence words and the aspect. BERT is a powerful pre-trained model which has achieved remarkable results in many NLP tasks. The architecture of BERT is a multi-layer bidirectional
Transformer Encoder \cite{Vaswani2017Attention}, which uses the self-attention mechanism to capture complex interaction and dependency between terms within a sequence. To leverage BERT to model the relationships between the sentence and the aspect, we pack the sentence and aspect together into a single sequence and then feed it into BERT, as shown in Figure \ref{network}. With this sentence-aspect concatenation, both the word-aspect associations and word-word dependencies are modeled interactively and simultaneously. 
With the contextual token representations $T_S=T_{[1:N]}\in\mathbb{R}^{N\times{H}}$ of the sentence, where $N$ is the sentence length and $H$ is the hidden size, we can then determine the start and end positions of the opinion snippet in the sentence.

\subsection{Soft-Selection Approach}
\label{bert-soft}
To fairly compare the performance of soft-selection approaches with hard-selection approaches, we use the same word-aspect fusion results $T_{S}$ from BERT. We implement the attention mechanism by
adopting the approach similar to the work \cite{Lin2017A}.
\begin{equation}
\begin{split}
\boldsymbol{\alpha} &= softmax(\boldsymbol{v_1}{tanh}({W_1}{T_{S}}^\mathrm{T}))  \\
\boldsymbol{g} &= \boldsymbol{\alpha}{T_S}
\end{split}
\end{equation}
where $\boldsymbol{v_1}\in\mathbb{R}^{H}$ and $W_1\in\mathbb{R}^{H\times{H}}$ are the parameters. The normalized attention weights $\boldsymbol{\alpha}$ are used to softly select words from the whole sentence and generate the final aspect-specific sentence representation $\boldsymbol{g}$. Then we make sentiment prediction as follows:

\begin{equation}
\label{eq:prediction}
\boldsymbol{\hat{y}} = softmax(W_2\boldsymbol{g} + \boldsymbol{b}) 
\end{equation}
where $W_2\in\mathbb{R}^{C\times{H}}$ and $\boldsymbol{b}\in\mathbb{R}^{C}$ are the weight matrix and bias vector respectively. $\boldsymbol{\hat{y}}$ is the probability distribution on $C$ polarities. The polarity with highest probability is selected as the prediction.

\subsection{Hard-Selection Approach}
\label{bert-hard}
Our proposed hard-selection approach determines the start and end positions of the opinion snippet and selects the words between these two positions for sentiment prediction. Since we do not have the ground-truth opinion snippet, but only the polarity of the corresponding aspect, we adopt reinforcement learning \cite{Williams1992Simple} to train our model. To make sure that the end position comes after the start position, we determine the start and end sequentially as a sequence training problem \cite{rennie2017self}. The parameters of the network, $\Theta$, define a policy $p_{\theta}$ and output an “action” that is the prediction of the position. For simplicity, we only generate two actions for determining the start and end positions respectively. After determining the start position, the ``state" is updated and then the end is conditioned on the start. 

Specifically, we define a start vector $\boldsymbol{s}\in\mathbb{R}^{H}$ and an end vector $\boldsymbol{e}\in\mathbb{R}^{H}$. Similar to the prior work \cite{Devlin2018BERT}, the probability of a word being the start of the opinion snippet is computed as a dot product between its contextual token representation and $\boldsymbol{s}$ followed by a softmax over all of the words of the sentence.

\begin{equation}
    \boldsymbol{\beta_l} = softmax(T_S\boldsymbol{s})
\end{equation}
We then sample the start position $l$ based on the multinomial distribution $\boldsymbol{\beta_l}$. To guarantee the end comes after the start, the end is sampled only in the right part of the sentence after the start. Therefore, the state is updated by slicing operation ${T_S}^r=T_S[l:]$. Same as the start position, the end position $r$ is also sampled based on the distribution $\boldsymbol{\beta_r}$: 
\begin{equation}
    \boldsymbol{\beta_r} = softmax(T_S^r\boldsymbol{e})
\end{equation}
Then we have the opinion snippet $T_O=T_S{[l:r]}$ to predict the sentiment polarity of the given aspect in the sentence. The probabilities of the start position at $l$ and the end position at $r$ are $p(l)=\boldsymbol{\beta_l}[l]$ and $p(r)=\boldsymbol{\beta_r}[r]$ respectively.

\subsubsection{Reward} 
After we get the opinion snippet $T_O$ by the sampling of the start and end positions, we compute the final representation $\boldsymbol{g_o}$ by the average of the opinion snippet,  $\boldsymbol{g_o}=avg(T_O)$. Then, equation \ref{eq:prediction} with different weights
is applied for computing the sentiment prediction $\boldsymbol{\hat{y_o}}$.
The cross-entropy loss function is employed for computing the reward.
\begin{equation}
R = -\sum\limits_{c}y^c\log{\boldsymbol{\hat{{y_o}^c}}}
\end{equation}
where $c$ is the index of the polarity class and $y$ is the ground truth.

\subsubsection{Self-Critical Training} 
In this paper, we use reinforcement learning to learn the start and end positions. The goal of training is to minimize the negative expected reward as shown below. 
\begin{equation}
L(\Theta) = -R\cdot p(l) \cdot p(r)
\end{equation}
where $\Theta$ is all the parameters in our architecture, which includes the base method BERT, the position selection parameters $\{\boldsymbol{s},\boldsymbol{e}\}$, and the parameters for sentiment prediction and then for reward calculation. Therefore, the \emph{state} in our method is the combination of the sentence and the aspect. For each state, the \emph{action} space is every position of the sentence.
\newline\indent
To reduce the variance of the gradient estimation, the reward is associated with a reference reward or baseline $R_b$ \cite{rennie2017self}. With the likelihood ratio trick, the objective function can be transformed as.
\begin{equation}
    L(\Theta) = -(R-R_b)\cdot (log(p(l))+log(p(r)))
\end{equation}
The baseline $R_b$ is computed based on the snippet determined by the baseline policy, which selects the start and end positions greedily by the $argmax$ operation on the $softmax$ results. As shown in Figure \ref{network}, the reward $R$ is calculated by sampling the snippet, while the baseline $R_b$ is computed by greedily selecting the snippet. Note that in the test stage, the snippet is determined by $argmax$ for inference.

\section{Experiments}
\label{sec:exp}
In this section, we compare our hard-selection model with various baselines. To assess the ability of alleviating the attention distraction, we further conduct experiments on a simulated multi-aspect dataset in which each sentence contains multiple aspects.
\subsection{Datasets}
We use the same datasets as the work by \newcite{tay2018learning}, which are already processed to token lists and released in Github\footnote{https://github.com/vanzytay/ABSA\_DevSplits}. The datasets are from SemEval 2014 task 4 \cite{Pontiki2014SemEval}, and SemEval 2015 task 12 \cite{semeval-2015}, respectively. For aspect term level sentiment classification task (denoted by T), we apply the Laptops and Restaurants datasets from SemEval 2014. For aspect category level sentiment prediction (denoted by C), we utilize the Restaurants dataset from SemEval 2014 and a composed dataset from both SemEval 2014 and SemEval 2015. The statistics of the datasets are shown in Table \ref{table-dataset}.

\begin{table}[t!]
\begin{center}
\setlength{\tabcolsep}{0.9mm}{
\begin{tabular} {|c|c|cccc|}
\hline
	Task &  Dataset &  \emph{All} &  \emph{P} & \emph{N} & \emph{Nu}\\
	\hline
		 T & Laptops Train & 1813 & 767 & 673 & 373\\
         T & Laptops Dev & 500 & 220 & 193 & 87\\
		 T & Laptops Test & 638 & 341 & 128 & 169 \\
     \hline
         T & Restaurants Train & 3102 & 685 & 1886 & 531 \\
         T & Restaurants Dev & 500 & 278 & 120 & 102\\
         T & Restaurants Test & 1120 & 728 & 196 & 196 \\
	\hline
         C & Restaurants Train & 3018 & 1873 & 712 & 433 \\
         C & Restaurants Dev & 500 & 306 & 127 & 67\\
         C & Restaurants Test & 973 & 657 & 222 & 94 \\
     \hline
         C & SE 14+15 Train & 3587 & 1069 & 2310 & 208 \\
         C & SE 14+15 Dev & 427 & 274 & 134 & 19\\
         C & SE 14+15 Test & 1011 & 455 & 496 & 60 \\
\hline
\end{tabular}}
\end{center}
\caption{\label{table-dataset} Dataset statistics. T and C denote the aspect-term and aspect-category tasks, respectively. \emph{P}, \emph{N}, and \emph{Nu} represent the numbers of instances with positive, negative and neutral polarities, and \emph{All} is the total number of instances.}	
\end{table}

\begin{table*}
\begin{center}
\setlength{\tabcolsep}{1.2mm}{
\begin{tabular}{|c|c|cccc|cccc|c|}
\hline
 &  & \multicolumn{4}{c|}{Term-Level} & \multicolumn{4}{c|}{Category-Level} &  \\
 &  & \multicolumn{2}{c}{Laptops} & \multicolumn{2}{c|}{Restaurants} & \multicolumn{2}{c}{Restaurants} & \multicolumn{2}{c|}{SemEval 14+15} &  \\
\hline
{\bf Model} & Aspect & 3-way & Binary & 3-way & Binary & 3-way & Binary & 3-way & Binary & Avg \\
\hline
LSTM & No & 61.75 & 78.25 & 67.94 & 82.03 & 73.38 & 79.97 & 75.96 & 79.92 & 74.90 \\
TD-LSTM & Yes & 62.38 & 79.31 & 69.73 & 84.41 & 79.97 & 75.96 & 79.92 & 74.90 & 75.63 \\
AT-LSTM & Yes & 65.83 & 78.25 & 74.37 & 84.74 & 77.90 & 84.87 & 76.16 & 81.28 & 77.93 \\
ATAE-LSTM & Yes & 60.34 & 74.20 & 70.71 & 84.52 & 77.80 & 83.85 & 74.08 & 78.96 & 75.56 \\
AF-LSTM(CORR) & Yes & 64.89 & 79.96 & 74.76 & 86.91 & 80.47 & 86.58 & 74.68 & 81.60 & 78.73 \\
AF-LSTM(CONV) & Yes & 68.81 & 83.58 & 75.44 & 87.78 & 81.29 & 87.26 & 78.44 & 81.49 & 80.51 \\
BERT-Original & Yes & 74.57 & 88.25 & 82.66 & {\bf 92.31} & {\bf 88.17} & 92.37 & 80.50 & 86.84 & 85.71\\
\hline
BERT-Soft & Yes & {\bf 74.92} & {\bf 90.41} & 82.68 & 91.98 & 87.05 & 91.92 & 80.02 & 86.75 & 85.72 \\
BERT-Hard & Yes & 74.10 & 89.55  & {\bf 83.91} & {\bf 92.31} & {\bf 88.17} & {\bf 93.39} & {\bf 81.09} & {\bf 87.89} & {\bf 86.30} \\
\hline
\end{tabular}}
\end{center}
\caption{Experimental results (accuracy \%)  on all the datasets. Models in the first part are baseline methods. The results in the first part (except BERT-Original) are obtained from the prior work \protect\cite{tay2018learning}. Avg column presents macro-averaged results across all the datasets.}
\label{table:final-result}
\end{table*}

\subsection{Implementation Details}
Our proposed models are implemented in PyTorch\footnote{https://github.com/huggingface/pytorch-pretrained-BERT}. We utilize the bert-base-uncased model, which contains 12 layers and the number of all parameters is 100M. The dimension $H$ is 768. The BERT model is initialized from the pre-trained model, other parameters are initialized by sampling from normal distribution $\mathcal{N}(0,0.02)$. In our experiments, the batch size is 32. The reported results are the testing scores that fine-tuning 7 epochs with learning rate 5e-5.

\subsection{Compared Models}
\begin{itemize}
\item {\bf LSTM}: it uses the average of all hidden states as the sentence representation for sentiment prediction. In this model, aspect information is not used.

\item {\bf TD-LSTM} \cite{tang2015target}: it employs two LSTMs and both of their outputs are applied to predict the sentiment polarity.

\item {\bf AT-LSTM} \cite{wang2016attention}: it utilizes the attention mechanism to produce an aspect-specific sentence representation. This method is a kind of soft-selection approach. 

\item {\bf ATAE-LSTM} \cite{wang2016attention}: it also uses the attention mechanism. The difference with AT-LSTM is that it concatenates the aspect embedding to each word embedding as the input to LSTM.

\item {\bf AF-LSTM(CORR)} \cite{tay2018learning}: it adopts circular correlation to capture the deep fusion between sentence words and the aspect, which can learn rich, higher-order relationships between words and the aspect.

\item {\bf AF-LSTM(CONV)} \cite{tay2018learning}: compared with AF-LSTM(CORR), this method applies circular convolution of vectors for performing word-aspect fusion to learn relationships between sentence words and the aspect.

\item {\bf BERT-Original}: it makes sentiment prediction by directly using the final hidden vector $C$ from BERT with the sentence-aspect pair as input.

\end{itemize}

\subsection{Our Models}
\begin{itemize}
\item {\bf BERT-Soft}: as described in Section \ref{bert-soft}, the contextual token representations from BERT are processed by self attention mechanism \cite{Lin2017A} and the attention-weighted sentence representation is utilized for sentiment classification.

\item {\bf BERT-Hard}: as described in Section \ref{bert-hard}, it takes the same input as BERT-Soft. It is called a hard-selection approach since it employs reinforcement learning techniques to explicitly select the opinion snippet corresponding to a particular aspect for sentiment prediction. 
\end{itemize}

\subsection{Experimental Results}
In this section, we evaluate the performance of our models by comparing them with various baseline models. Experimental results are illustrated in Table \ref{table:final-result}, in which \emph{3-way} represents 3-class sentiment classification (\emph{positive}, \emph{negative} and \emph{neutral}) and \emph{Binary} denotes binary sentiment prediction (\emph{positive} and \emph{negative}). The best score of each column is marked in bold.

Firstly, we observe that BERT-Original, BERT-Soft, and BERT-Hard outperform all soft attention baselines (in the first part of Table \ref{table:final-result}),  which demonstrates the effectiveness of fine-tuning the pre-trained model on the aspect-based sentiment classification task. Particularly, BERT-Original outperforms AF-LSTM(CONV) by 2.63\%$\sim$9.57\%, BERT-Soft outperforms AF-LSTM(CONV) by 2.01\%$\sim$9.60\% and BERT-Hard improves AF-LSTM(CONV) by 3.38\%$\sim$11.23\% in terms of accuracy. Considering the average score across eight settings, BERT-Original outperforms AF-LSTM(CONV) by 6.46\%, BERT-Soft outperforms AF-LSTM(CONV) by 6.47\% and BERT-Hard outperforms AF-LSTM(CONV) by 7.19\% respectively.

Secondly, we compare the performance of three BERT-related methods. The performance of BERT-Original and BERT-Soft are similar by comparing their average scores. The reason may be that the original BERT has already modeled the deep relationships between the sentence and the aspect. BERT-Original can be thought of as a kind of soft-selection approach as BERT-Soft. We also observe that the snippet selection by reinforcement learning improves the performance over soft-selection approaches in almost all settings. However, the improvement of BERT-Hard over BERT-Soft is marginal. The average score of BERT-Hard is better than BERT-Soft by 0.68\%. The improvement percentages are between 0.36\% and 1.49\%, while on the Laptop dataset, the performance of BERT-Hard is slightly weaker than BERT-Soft. The main reason is that the datasets only contain a small portion of multi-aspect sentences with different polarities. The distraction of attention will not impact the sentiment prediction much in single-aspect sentences or multi-aspect sentences with the same polarities.

\subsection{Experimental Results on Multi-Aspect Sentences}
On the one hand, the attention distraction issue becomes worse in multi-aspect sentences. In addition to noisy and misleading words, the attention is also prone to be distracted by opinion words from other aspects of the sentence. On the other hand, the attention distraction impacts the performance of sentiment prediction more in multi-aspect sentences than in single-aspect sentences. Hence, we evaluate the performance of our models on a test dataset with only multi-aspect sentences. 

A multi-aspect sentence can be categorized by two dimensions: the \emph{Number} of aspects and the \emph{Polarity} dimension which indicates whether the sentiment polarities of all aspects are the same or not. In the dimension of \emph{Number}, we categorize the multi-aspect sentences as \emph{2-3} and \emph{More}. \emph{2-3} refers to the sentences with two or three aspects while \emph{More} refers to the sentences with more than three aspects. The statistics in the original dataset shows that there are much more sentences with \emph{2-3} aspects than those with \emph{More} aspects. In the dimension \emph{Polarity}, the multi-aspect sentences can be categorized into \emph{Same} and \emph{Diff}. \emph{Same} indicates that all aspects in the sentence have the same sentiment polarity. \emph{Diff} indicates that the aspects have different polarities. 

{\bf Multi-aspect test set.}\quad To evaluate the performance of our models on multi-aspect sentences, we construct a new multi-aspect test set by selecting all multi-aspect sentences from the original training, development, and test sets of the Restaurants term-level task. The details are shown in Table \ref{table-testmulti}. 

\begin{table} [t!]
\begin{center}
\setlength{\tabcolsep}{0.7mm}{
\begin{tabular} {|c|ccc|ccc|c|}
\hline
		  \multirow{2}{*}{Type} & \multicolumn{3}{c|}{\emph{Same}} & \multicolumn{3}{c|}{\emph{Diff}} & \multirow{2}{*}{Total}\\
		  \cline{2-7}
		       &  \emph{2-3} & \emph{More} & Total  & \emph{2-3}  & \emph{More}  & Total &  \\
		  \hline
          Number & 1665 & 352 & 2017 & 655 & 327 & 982 & 2999\\
\hline
\end{tabular}}
\end{center}
\caption{\label{table-testmulti}Distribution of the multi-aspect test set. Around 67\% of the multi-aspect sentences belong to the \emph{Same} category.}
\end{table}

\begin{table} [t!]
\begin{center}
\setlength{\tabcolsep}{0.3mm}{
\begin{tabular} {|c|c|ccc|ccc|c|}
\hline
      \multicolumn{8}{|c|}{Constructed Multi-Aspect Training Set} & Total \\
      \hline
      \hline
	  \multicolumn{2}{|c}{\multirow{2}{*}{Single}} & \multicolumn{2}{|c|}{\emph{P}} & \multicolumn{2}{c|}{\emph{N}} & \multicolumn{2}{c|}{\emph{Nu}} &  \multirow{2}{*}{891}\\
	  \cline{3-8}
	     \multicolumn{2}{|c|}{}  &  \multicolumn{2}{c|}{297}  & \multicolumn{2}{c|}{297}  & \multicolumn{2}{c|}{297} &   \\
	  \hline
	  \hline
          \multicolumn{2}{|c|}{} & \multicolumn{3}{c|}{\emph{Same}} & \multicolumn{3}{c|}{\emph{Diff}} & \\
      \hline
          \multirow{4}{*}{Multi} & \multirow{2}{*}{2-asp} & \emph{2P} & \emph{2N} & \emph{2Nu} & \emph{PN} & \emph{PNu} & \emph{NNu} & \multirow{4}{*}{3600}\\
          & & 300 & 300 & 300 & 300 & 300 & 300 & \\
      \cline{2-8}
      & \multirow{2}{*}{3-asp} & \emph{3P} & \emph{3N} & \emph{3Nu} & \emph{2P1N} & \emph{1P2N} & \emph{PNNu} & \\
          & & 300 & 300 & 300 & 300 & 300 & 300 & \\
\hline
\end{tabular}}
\end{center}
\caption{\label{table-trainnumber} Distribution of the multi-aspect training set. 2-asp and 3-asp indicate that the sentence contains two or three aspects respectively. Each multi-aspect sentence is  categorized as \emph{Same} or \emph{Diff}.}	
\end{table}

\begin{figure*}
\centering
	\includegraphics[width=0.99\textwidth]{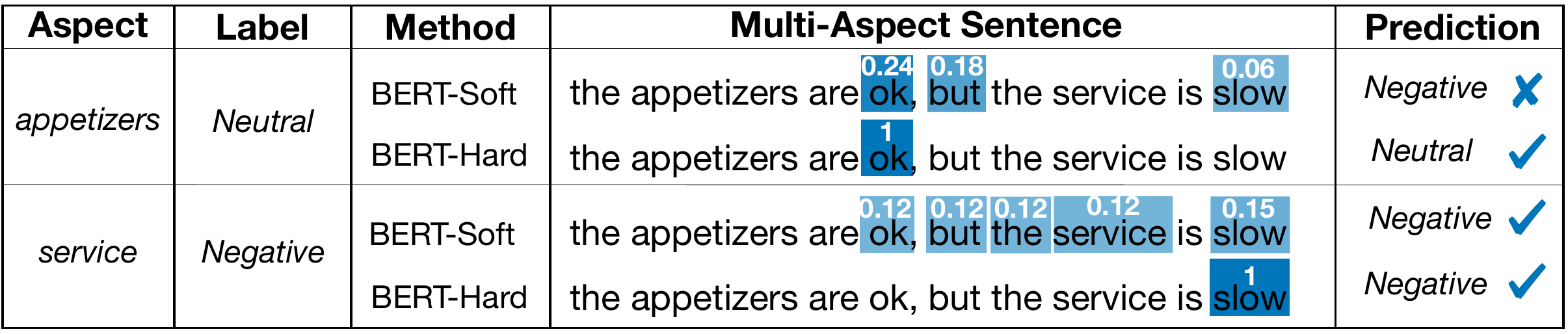}
    \caption{Visualization. The attention weights are visualized for BERT-Soft, and the selected opinion snippets are marked for BERT-Hard. The correctness of the predicted results is also marked.}
    \label{figure:case}
\end{figure*}

{\bf Multi-aspect training set.}\quad Since we use all multi-aspect sentences for testing, we need to generate some ``virtual'' multi-aspect sentences for training. The simulated multi-aspect training set includes the original single-aspect sentences and the newly constructed multi-aspect sentences, which are generated by concatenating multiple single-aspect sentences with different aspects. We keep the balance of each subtype in the new training set (see Table \ref{table-trainnumber}). The number of \emph{Neutral} sentences is the least among three sentiment polarities in all single-aspect sentences. We randomly select the same number of \emph{Positive} and \emph{Negative} sentences. Then we construct multi-aspect sentences by combining single-aspect sentences in different combinations of polarities. The naming for different combinations is simple. For example, \emph{2P-1N} indicates that the sentence has two positive aspects and one negative aspect, and \emph{P-N-Nu} means that the three aspects in the sentence are positive, negative, and neutral respectively. For simplicity, we only construct 2-asp and 3-asp sentences which are also the majority in the original dataset.

{\bf Results and Discussions.}\quad 
The results on different types of multi-aspect sentences are shown in  Table \ref{table-trainmulti}. The performance of BERT-Hard is better than BERT-Original and BERT-Soft over all types of multi-aspect sentences. BERT-Hard outperforms BERT-Soft by 2.11\% when the aspects have the same sentiment polarities. For multi-aspect sentences with different polarities, the improvements are more significant. BERT-Hard outperforms BERT-Soft by 7.65\% in total of \emph{Diff}. The improvements are 5.07\% and 12.83\% for the types \emph{2-3} and \emph{More} respectively, which demonstrates the ability of our model on handling sentences with \emph{More} aspects. Particularly, BERT-Soft has the poorest performance on the subset \emph{Diff} among the three methods, which proves that soft attention is more likely to cause attention distraction.

\begin{table} [t!]
\begin{center}
\setlength{\tabcolsep}{0.9mm}{
\begin{tabular} {|c|c|ccc|c|}
\hline
		  \multirow{2}{*}{Type} & \multirow{2}{*}{\emph{Same}} & \multicolumn{3}{c|}{\emph{Diff}} & \multirow{2}{*}{Total}\\
		  \cline{3-5}
		       &      & \emph{2-3}  & \emph{More}  & Total &  \\
     \hline
          BERT-Original & 73.33 & 57.10 & 60.86 & 58.35 & 68.42 \\
          BERT-Soft & 75.31 & 57.25 & 57.19 & 57.23 & 69.39 \\
          BERT-Hard & {\bf 76.90} & {\bf 60.15} & {\bf 64.53} & {\bf 61.61} & {\bf 71.89}\\
\hline
\end{tabular}}
\end{center}
\caption{\label{table-trainmulti} Experimental results (accuracy \%) on multi-aspect sentences. The performance of the 3-way classification on the multi-aspect test set is reported.}	
\end{table}

Intuitively, when multiple aspects in the sentence have the same sentiment polarities, even the attention is distracted to other opinion words of other aspects, it can still predict correctly to some extent. In such sentences, the impact of the attention distraction is not obvious and difficult to detect. However, when the aspects have different sentiment polarities, the attention distraction will lead to catastrophic error prediction, which will obviously decrease the classification accuracy. As shown in Table \ref{table-trainmulti}, the accuracy of \emph{Diff} is much worse than \emph{Same} for all three methods. It means that the type of \emph{Diff} is difficult to handle. Even though, the significant improvement proves that our hard-selection method can alleviate the attention distraction to a certain extent. For soft-selection methods, the attention distraction is inevitable due to their way in calculating the attention weights for every single word. The noisy or irrelevant words could seize more attention weights than the ground truth opinion words. Our method considers the opinion snippet as a consecutive whole, which is more resistant to attention distraction. 

\subsection{Visualization}
In this section, we visualize the attention weights for BERT-Soft and opinion snippets for BERT-Hard. As demonstrated in Figure \ref{figure:case}, the multi-aspect sentence \emph{``the appetizers are OK, but the service is slow''} belongs to the category \emph{Diff}. Firstly, the attention weights of BERT-Soft scatter among the whole sentence and could attend to irrelevant words. For the aspect \emph{service}, BERT-Soft attends to the word \emph{``ok''} with relatively high score though it does not describe the aspect \emph{service}. This problem also exists for the aspect \emph{appetizers}. Furthermore, the attention distraction could cause error prediction. For the aspect \emph{appetizers}, \emph{``but''} and \emph{``slow''} gain high attention scores and cause the wrong sentiment prediction \emph{Negative}.

Secondly, our proposed method BERT-Hard can detect the opinion snippet for a given aspect. As illustrated in Figure \ref{figure:case}, the opinion snippets are selected by BERT-Hard accurately. In the sentence \emph{``the appetizers are ok, but the service is slow''}, BERT-Hard can exactly locate the opinion snippets \emph{``ok''} and \emph{``slow''} for the aspect \emph{appetizers} and \emph{service} respectively. 

At last, we enumerate some opinion snippets detected by BERT-Hard in Table \ref{table-snippets}. Our method can precisely detect snippets even for latent opinion expression and alleviate the influence of noisy words. For instance, \emph{``cannot be beat for the quality''} is hard to predict using soft attention because the sentiment polarity is transformed by the negative word \emph{``cannot''}. Our method can select the whole snippet without bias to any word and in this way the attention distraction can be alleviated. We also list some inaccurate snippets in Table \ref{table-snippets-f}. Some meaningless words around the true snippet are included, such as \emph{``are''}, \emph{``and''} and \emph{``at''}. These words do not affect the final prediction. A possible explanation to these inaccurate words is that the true snippets are unlabeled and our method predicts them only by the supervisory signal from sentiment labels. 

\begin{table} [t!]
\small
\begin{center}
\setlength{\tabcolsep}{1.4mm}{
\begin{tabular} {c|c}
\hline
		  Positive Snippets & Negative Snippets \\
     \hline
         very good prompt attentive & not great bland \\
         beautifully presented &  can not eat this well \\
         extremely tasty &  unbearable conversation  \\
         as interesting as possible &  no idea how to use\\
         cool and soothing &  would never go there\\
         impressed by & not above ordinary\\
         cannot be beat for the quality & not good\\
\hline
\end{tabular}}
\end{center}
\caption{\label{table-snippets} Examples of accurate opinion snippets detected by BERT-Hard.}	
\end{table}

\begin{table} [t!]
\small
\begin{center}
\setlength{\tabcolsep}{1.4mm}{
\begin{tabular} {c | c}
\hline
		  \multicolumn{2}{c}{Inaccurate Snippets} \\
     \hline
         are very large and  & and even greater food \\
         are not terrible  & tasty treat at \\
         everyone who works & the money and said \\
\hline
\end{tabular}}
\end{center}
\caption{\label{table-snippets-f} Examples of inaccurate opinion snippets detected by BERT-Hard.}	
\end{table}

\section{Conclusion}
\label{sec:conclude}
In this paper, we propose a {\bf hard-selection} approach for aspect-based sentiment analysis, which determines the start and end positions of the opinion snippet for a given input aspect. The deep associations between the sentence and aspect, and the long-term dependencies within the sentence are taken into consideration by leveraging the pre-trained BERT model. With the hard selection of the opinion snippet, our approach can alleviate the attention distraction problem of traditional attention-based soft-selection methods. Experimental results demonstrate the effectiveness of our method. Especially, our hard-selection approach outperforms soft-selection approaches significantly when handling multi-aspect sentences with different sentiment polarities.

\section{Acknowledgement}
This work is supported by National Science and Technology Major Project, China (Grant No. 2018YFB0204304).

\bibliography{conll-2019}

\begin{thebibliography}{28}
\expandafter\ifx\csname natexlab\endcsname\relax\def\natexlab#1{#1}\fi

\bibitem[{Bahdanau et~al.(2014)Bahdanau, Cho, and Bengio}]{Bahdanau2014Neural}
Dzmitry Bahdanau, Kyunghyun Cho, and Yoshua Bengio. 2014.
\newblock Neural machine translation by jointly learning to align and
  translate.
\newblock \emph{Computer Science}.

\bibitem[{Boiy and Moens(2009)}]{boiy2009machine}
Erik Boiy and Marie-Francine Moens. 2009.
\newblock A machine learning approach to sentiment analysis in multilingual web
  texts.
\newblock \emph{Information retrieval}, 12(5):526--558.

\bibitem[{Chen et~al.(2017)Chen, Sun, Bing, and Yang}]{chen2017recurrent}
Peng Chen, Zhongqian Sun, Lidong Bing, and Wei Yang. 2017.
\newblock Recurrent attention network on memory for aspect sentiment analysis.
\newblock In \emph{Proceedings of the 2017 conference on empirical methods in
  natural language processing (EMNLP)}, pages 452--461.

\bibitem[{Cheng et~al.(2017)Cheng, Zhao, Zhang, King, Zhang, and
  Wang}]{cheng2017aspect}
Jiajun Cheng, Shenglin Zhao, Jiani Zhang, Irwin King, Xin Zhang, and Hui Wang.
  2017.
\newblock Aspect-level sentiment classification with heat (hierarchical
  attention) network.
\newblock In \emph{Proceedings of the 2017 ACM on Conference on Information and
  Knowledge Management (CIKM)}, pages 97--106. ACM.

\bibitem[{Devlin et~al.(2018)Devlin, Chang, Lee, and
  Toutanova}]{Devlin2018BERT}
Jacob Devlin, Ming~Wei Chang, Kenton Lee, and Kristina Toutanova. 2018.
\newblock Bert: Pre-training of deep bidirectional transformers for language
  understanding.

\bibitem[{Ding et~al.(2009)Ding, Liu, and Zhang}]{ding2009entity}
Xiaowen Ding, Bing Liu, and Lei Zhang. 2009.
\newblock Entity discovery and assignment for opinion mining applications.
\newblock In \emph{Proceedings of the 15th ACM international conference on
  Knowledge discovery and data mining (SIGKDD)}, pages 1125--1134.

\bibitem[{Fan et~al.(2018)Fan, Feng, and Zhao}]{fan2018multi}
Feifan Fan, Yansong Feng, and Dongyan Zhao. 2018.
\newblock Multi-grained attention network for aspect-level sentiment
  classification.
\newblock In \emph{Proceedings of the 2018 Conference on Empirical Methods in
  Natural Language Processing (EMNLP)}, pages 3433--3442.

\bibitem[{Hazarika et~al.(2018)Hazarika, Poria, Vij, Krishnamurthy, Cambria,
  and Zimmermann}]{hazarika2018modeling}
Devamanyu Hazarika, Soujanya Poria, Prateek Vij, Gangeshwar Krishnamurthy, Erik
  Cambria, and Roger Zimmermann. 2018.
\newblock Modeling inter-aspect dependencies for aspect-based sentiment
  analysis.
\newblock In \emph{Proceedings of the 2018 Conference of the North American
  Chapter of the Association for Computational Linguistics: Human Language
  Technologies (NAACL-HLT)}, pages 266--270.

\bibitem[{Jiang et~al.(2011)Jiang, Yu, Zhou, Liu, and Zhao}]{jiang2011target}
Long Jiang, Mo~Yu, Ming Zhou, Xiaohua Liu, and Tiejun Zhao. 2011.
\newblock Target-dependent twitter sentiment classification.
\newblock In \emph{Proceedings of the 49th Annual Meeting of the Association
  for Computational Linguistics (ACL)}, pages 151--160.

\bibitem[{Li et~al.(2018{\natexlab{a}})Li, Liu, and Zhou}]{li2018hierarchical}
Lishuang Li, Yang Liu, and AnQiao Zhou. 2018{\natexlab{a}}.
\newblock Hierarchical attention based position-aware network for aspect-level
  sentiment analysis.
\newblock In \emph{Proceedings of the 22nd Conference on Computational Natural
  Language Learning (CoNLL)}, pages 181--189.

\bibitem[{Li et~al.(2018{\natexlab{b}})Li, Bing, Lam, and Shi}]{li2018acl}
Xin Li, Lidong Bing, Wai Lam, and Bei Shi. 2018{\natexlab{b}}.
\newblock Transformation networks for target-oriented sentiment classification.
\newblock In \emph{Proceedings of the 56th Annual Meeting of the Association
  for Computational Linguistics (ACL)}.

\bibitem[{Lin et~al.(2017)Lin, Feng, Santos, Mo, Bing, Zhou, and
  Bengio}]{Lin2017A}
Zhouhan Lin, Minwei Feng, Cicero Nogueira~Dos Santos, Yu~Mo, Xiang Bing, Bowen
  Zhou, and Yoshua Bengio. 2017.
\newblock A structured self-attentive sentence embedding.
\newblock In \emph{The 5th International Conference on Learning Representations
  (ICLR)}.

\bibitem[{Liu(2012)}]{liu2012sentiment}
Bing Liu. 2012.
\newblock Sentiment analysis and opinion mining.
\newblock \emph{Synthesis lectures on human language technologies},
  5(1):1--167.

\bibitem[{Ma et~al.(2017)Ma, Li, Zhang, Wang, Ma, Li, Zhang, and
  Wang}]{Ma2017Interactive}
Dehong Ma, Sujian Li, Xiaodong Zhang, Houfeng Wang, Dehong Ma, Sujian Li,
  Xiaodong Zhang, and Houfeng Wang. 2017.
\newblock Interactive attention networks for aspect-level sentiment
  classification.
\newblock In \emph{Twenty-Sixth International Joint Conference on Artificial
  Intelligence (IJCAI)}, pages 4068--4074.

\bibitem[{Majumder et~al.(2018)Majumder, Poria, Gelbukh, Akhtar, Cambria, and
  Ekbal}]{majumder2018iarm}
Navonil Majumder, Soujanya Poria, Alexander Gelbukh, Md~Shad Akhtar, Erik
  Cambria, and Asif Ekbal. 2018.
\newblock Iarm: Inter-aspect relation modeling with memory networks in
  aspect-based sentiment analysis.
\newblock In \emph{Proceedings of the 2018 Conference on Empirical Methods in
  Natural Language Processing (EMNLP)}, pages 3402--3411.

\bibitem[{Pang and Lee(2008)}]{Pang:2008:OMS:1454711.1454712}
Bo~Pang and Lillian Lee. 2008.
\newblock \href {https://doi.org/10.1561/1500000011} {Opinion mining and
  sentiment analysis}.
\newblock \emph{Found. Trends Inf. Retr.}, 2(1-2):1--135.

\bibitem[{Pontiki et~al.(2015)Pontiki, Galanis, Papageorgiou, Manandhar, and
  Androutsopoulos}]{semeval-2015}
Maria Pontiki, Dimitris Galanis, Haris Papageorgiou, Suresh Manandhar, and Ion
  Androutsopoulos. 2015.
\newblock {S}em{E}val-2015 task 12: Aspect based sentiment analysis.
\newblock In \emph{{S}em{E}val 2015}.

\bibitem[{Pontiki et~al.(2014)Pontiki, Galanis, Pavlopoulos, Papageorgiou,
  Androutsopoulos, and Manandhar}]{Pontiki2014SemEval}
Maria Pontiki, Dimitris Galanis, John Pavlopoulos, Harris Papageorgiou, Ion
  Androutsopoulos, and Suresh Manandhar. 2014.
\newblock Semeval-2014 task 4: Aspect based sentiment analysis.
\newblock \emph{(SemEval 2014)}, pages 27--35.

\bibitem[{Rennie et~al.(2017)Rennie, Marcheret, Mroueh, Ross, and
  Goel}]{rennie2017self}
Steven~J Rennie, Etienne Marcheret, Youssef Mroueh, Jerret Ross, and Vaibhava
  Goel. 2017.
\newblock Self-critical sequence training for image captioning.
\newblock In \emph{IEEE Conference on Computer Vision and Pattern Recognition
  (CVPR)}, pages 7008--7024.

\bibitem[{Tang et~al.(2015)Tang, Qin, Feng, and Liu}]{tang2015target}
Duyu Tang, Bing Qin, Xiaocheng Feng, and Ting Liu. 2015.
\newblock Target-dependent sentiment classification with long short term
  memory.
\newblock \emph{CoRR, abs/1512.01100}.

\bibitem[{Tang et~al.(2016)Tang, Qin, and Liu}]{tang2016aspect}
Duyu Tang, Bing Qin, and Ting Liu. 2016.
\newblock Aspect level sentiment classification with deep memory network.
\newblock \emph{Proceedings of the 2016 conference on empirical methods in
  natural language processing (EMNLP)}.

\bibitem[{Tay et~al.(2018)Tay, Luu, and Hui}]{tay2018learning}
Yi~Tay, Anh~Tuan Luu, and Siu~Cheung Hui. 2018.
\newblock Learning to attend via word-aspect associative fusion for
  aspect-based sentiment analysis.
\newblock In \emph{The Thirty-Second Conference on the Association for the
  Advance of Artificial Intelligence (AAAI)}.

\bibitem[{Vaswani et~al.(2017)Vaswani, Shazeer, Parmar, Uszkoreit, Jones,
  Gomez, Kaiser, and Polosukhin}]{Vaswani2017Attention}
Ashish Vaswani, Noam Shazeer, Niki Parmar, Jakob Uszkoreit, Llion Jones,
  Aidan~N Gomez, {\L}ukasz Kaiser, and Illia Polosukhin. 2017.
\newblock Attention is all you need.
\newblock In \emph{Advances in neural information processing systems (NIPS)},
  pages 5998--6008.

\bibitem[{Wang and Lu(2018)}]{wang2018learning}
Bailin Wang and Wei Lu. 2018.
\newblock Learning latent opinions for aspect-level sentiment classification.
\newblock In \emph{The Thirty-Second Conference on the Association for the
  Advance of Artificial Intelligence (AAAI)}.

\bibitem[{Wang et~al.(2018{\natexlab{a}})Wang, Li, Li, Kang, Zhang, Si, and
  Zhou}]{wang2018clause}
Jingjing Wang, Jie Li, Shoushan Li, Yangyang Kang, Min Zhang, Luo Si, and
  Guodong Zhou. 2018{\natexlab{a}}.
\newblock Aspect sentiment classification with both word-level and clause-level
  attention networks.
\newblock In \emph{Twenty-Seventh International Joint Conference on Artificial
  Intelligence (IJCAI)}.

\bibitem[{Wang et~al.(2018{\natexlab{b}})Wang, Mazumder, Liu, Zhou, and
  Chang}]{wang2018target}
Shuai Wang, Sahisnu Mazumder, Bing Liu, Mianwei Zhou, and Yi~Chang.
  2018{\natexlab{b}}.
\newblock Target-sensitive memory networks for aspect sentiment classification.
\newblock In \emph{Proceedings of the 56th Annual Meeting of the Association
  for Computational Linguistics (ACL)}, pages 957--967.

\bibitem[{Wang et~al.(2016)Wang, Huang, Zhao et~al.}]{wang2016attention}
Yequan Wang, Minlie Huang, Li~Zhao, et~al. 2016.
\newblock Attention-based lstm for aspect-level sentiment classification.
\newblock In \emph{Proceedings of the 2016 conference on empirical methods in
  natural language processing (EMNLP)}, pages 606--615.

\bibitem[{Williams(1992)}]{Williams1992Simple}
Ronald~J. Williams. 1992.
\newblock Simple statistical gradient-following algorithms for connectionist
  reinforcement learning.
\newblock \emph{Machine Learning}, 8(3-4):229--256.

\end{thebibliography}
\bibliographystyle{acl_natbib}

\end{document}